\documentclass[reprint,NumberedRefs]{JASA}

\usepackage{siunitx}

\begin{document}

\title{Physics-informed neural operators for the \textit{in situ} characterization of locally reacting sound absorbers}

\author{Jonas M. Schmid}
\email{jonas.m.schmid@tum.de}
\affiliation{Chair of Vibroacoustics of Vehicles and Machines, Technical University of Munich, Garching, 85748, Germany}
\author{Johannes D. Schmid}
\affiliation{Chair of Vibroacoustics of Vehicles and Machines, Technical University of Munich, Garching, 85748, Germany}
\author{Martin Eser}
\affiliation{Independent Researcher, Munich, 81377, Germany}
\author{Steffen Marburg}
\affiliation{Chair of Vibroacoustics of Vehicles and Machines, Technical University of Munich, Garching, 85748, Germany}

\date{\today}

\begin{abstract}
Accurate knowledge of acoustic surface admittance or impedance is essential for reliable wave-based simulations, yet its \textit{in situ} estimation remains challenging due to noise, model inaccuracies, and restrictive assumptions of conventional methods. This work presents a physics-informed neural operator approach for estimating frequency-dependent surface admittance directly from near-field measurements of sound pressure and particle velocity. A deep operator network is employed to learn the mapping from measurement data, spatial coordinates, and frequency to acoustic field quantities, while simultaneously inferring a globally consistent surface admittance spectrum without requiring an explicit forward model. The governing acoustic relations, including the Helmholtz equation, the linearized momentum equation, and Robin boundary conditions, are embedded into the training process as physics-based regularization, enabling physically consistent and noise-robust predictions while avoiding frequency-wise inversion. The method is validated using synthetically generated data from a simulation model for two planar porous absorbers under semi free-field conditions across a broad frequency range. Results demonstrate accurate reconstruction of both real and imaginary admittance components and reliable prediction of acoustic field quantities. Parameter studies confirm improved robustness to noise and sparse sampling compared to purely data-driven approaches, highlighting the potential of physics-informed neural operators for \textit{in situ} acoustic material characterization.
\end{abstract}


\maketitle


\section{\label{sec:Introduction} Introduction}
Advances in computational resources and numerical methods have established wave-based simulation methods such as the finite element method (FEM) and the boundary element method (BEM) as standard tools for predicting acoustic fields in enclosed environments. However, their accuracy critically depends on an exact representation of all acoustically interacting boundary surfaces, while uncertainties in material properties remain a major source of modeling error~\cite{Vorlander.2013, Thydal.2021}. Within wave-based formulations, boundary behavior is described by the complex-valued surface admittance $Y$ or its reciprocal, the surface impedance $Z$~\cite{Marburg.2008}. These quantities encode both amplitude and phase relations between acoustic pressure and particle velocity at the boundary and are therefore essential for physically consistent simulations. While the real part of the admittance represents dissipative losses, the imaginary part accounts for reactive energy storage~\cite{Marburg.2011}. The use of admittance-type boundary conditions further implies local reaction, i.e. waves inside the material solely propagate along the surface-normal direction.

Standardized methods for acoustic material characterization exist, but they are based on idealized assumptions that limit their applicability. The impedance tube method (ISO~10534-2~\cite{ISO105342.1998}) is restricted to normal incidence and is sensitive to mounting conditions~\cite{Horoshenkov.2007, Eser.2025}. The reverberation room method (ISO~354~\cite{ISO354.2003}) assumes a perfectly diffuse sound field, which is rarely fulfilled at low frequencies~\cite{Nolan.2018}. Moreover, the resulting absorption coefficient provides only magnitude information and neglects phase effects, which are crucial for accurately describing wave phenomena. This limitation becomes critical in small rooms and in the low-frequency regime, where modal behavior dominates the sound field~\cite{CardosoSoares.2022}. However, comprehensive datasets of complex-valued surface admittance or impedance remain limited~\cite{Fratoni.2025}.

To address these limitations, \textit{in situ} or free-field measurement techniques have gained increasing attention as non-destructive approaches for characterizing acoustic materials under realistic conditions~\cite{Brandao.2015}. These methods typically aim to ensure well-defined sound incidence, either through measurements in (semi-)anechoic environments~\cite{Brandao.2011} or by applying time-gating techniques to suppress room reflections~\cite{Brandao.2025}. Most approaches are formulated as inverse problems, where sound pressure and/or particle velocity measurements in the near field of the sample are used to infer boundary properties based on physical wave propagation models~\cite{Tamura.1990}. Data acquisition can be performed sequentially or using microphone arrays arranged in different configurations~\cite{Richard.2019b}. As an alternative or complement, pressure-particle velocity (PU) probes enable simultaneous measurements of pressure and particle velocity at a single point close to the surface~\cite{BrieredelaHosserayeBaltazar.2022, Brandao.2012}. The sound field is typically represented as a superposition of elementary wave components, such as plane waves~\cite{Nolan.2020, Nolan.2026}, spherical waves~\cite{Richard.2017, Alkmim.2021}, or complex image sources~\cite{Brandao.2025, Eser.2021}. Estimating their amplitudes leads to an inverse problem that is usually ill-posed and highly sensitive to noise, modeling errors, and sparse measurement data. Regularization techniques (e.g., Tikhonov regularization~\cite{Ottink.2016}, sparsity-promoting approaches~\cite{Shen.2024}), and Bayesian approaches~\cite{Xiang.2020, Schmid.2021} are therefore employed to overcome this ill-posedness. For instance, Eser et al.~\cite{Eser.2023} have introduced a Bayesian approach based on sequential frequency transfer to estimate frequency-dependent admittance with quantified uncertainties. More recently, data-driven methods have been proposed to compensate for modeling inaccuracies such as edge diffraction effects caused by finite sample dimensions~\cite{Emmerich.2025, MullerGiebeler.2024, Zea.2023}. 

Despite these developments, the accuracy of model-based approaches remains strongly dependent on the fidelity of the underlying wave propagation model. Simplifying assumptions, introduced to render the inverse problem tractable, can lead to systematic errors in the estimated material parameters. A prominent example is the assumption of spherical wave reflection at an infinite, locally reacting planar surface, which has been shown to introduce significant biases~\cite{Brandao.2011}. Such discrepancies become particularly pronounced at low frequencies, where edge diffraction effects cannot be neglected~\cite{Brandao.2022}.

An alternative class of \textit{in situ} methods employs discretization-based forward models, such as FEM or BEM, to compute the acoustic field for given boundary conditions~\cite{Anderssohn.2007}. The inverse problem is then solved using iterative optimization schemes~\cite{Luo.2020, Prinn.2021} or Bayesian inference~\cite{Schmid.2023}. Recent work has demonstrated the feasibility of such approaches for complex geometries with uncertainty quantification~\cite{Schmid.2023, Wulbusch.2024}. However, these methods are computationally demanding due to repeated forward simulations and typically require frequency-wise inversion. Schmid et al.~\cite{Schmid.2026} have recently proposed a simulation-based inference approach using neural networks to directly learn frequency-dependent surface impedances based on previously computed FEM simulations~\cite{Schmid.2026}. Nevertheless, this method still relies on an accurate forward model and remains sensitive to model–data mismatch~\cite{Wulbusch.2024}.

Physics-informed neural networks (PINNs) have recently emerged as a powerful approach that incorporates governing physical equations directly into the learning process~\cite{Raissi.2019}. This enables physically consistent predictions even in the presence of noisy or sparse data~\cite{Karniadakis.2021}. In acoustics, PINNs have been applied to sound field prediction~\cite{BorrelJensen.2021}, radiation problems~\cite{Schmid.2025}, and sound field reconstruction~\cite{Karakonstantis.2024}. Recent studies have demonstrated their potential for inferring acoustic boundary properties directly from pressure measurements~\cite{Schmid.2024, Xia.2024}. However, these approaches require separate model training for each frequency. A very recent preprint by Xia et al.~\cite{Xia.2026} proposes a physics-informed neural field approach that reconstructs the acoustic field from sparse pressure data and derives impedance on the surface from pressure gradients. While eliminating the need for particle velocity measurements, this approach increases sensitivity to noise and reduces accuracy at low frequencies due to inaccuracies in gradient estimation. A fundamental limitation of PINNs lies in their formulation as solvers for a single instance of a parameterized partial differential equation (PDE). As a result, any change in input parameters, such as excitation frequency, requires retraining of the model. This restricts their efficiency and scalability for parametric problems, particularly for frequency-dependent phenomena that are omnipresent in acoustics.

Neural operators address this limitation by learning mappings between function spaces rather than solving PDEs for individual parameter instances~\cite{Kovachki.2023, Azizzadenesheli.2024}. Once trained, they enable rapid predictions of solution fields for previously unseen inputs, such as new frequencies or spatial coordinates, at negligible computational cost. Neural operators have recently been applied to various problems in acoustics and structural dynamics, including sound field prediction~\cite{BorrelJensen.2024, Middleton.2025}, acoustic scattering~\cite{Nair.2025}, and structural intensity estimation~\cite{Schmid.2026b}.

In this work, a physics-informed deep operator network (DeepONet) is proposed to estimate frequency-dependent acoustic surface admittance directly from near-field measurements of sound pressure and normal particle velocity. By explicitly incorporating frequency as an input variable, the method estimates the full surface admittance spectrum within a single model, eliminating the need for frequency-wise inversion. At the same time, it avoids explicit analytical or numerical forward models by learning the mapping from measurements to acoustic field quantities and boundary parameters directly, thereby reducing computational cost and mitigating model-induced errors. The governing acoustic relations are directly embedded into the training process, providing physics-based regularization that enhances robustness and physical consistency, offering a promising alternative to conventional methods.

\subsection{\label{subsec:Methodological overview} Methodological overview}
\begin{figure*}[tbh]
		\centering
		\includegraphics[width = 0.9\textwidth]{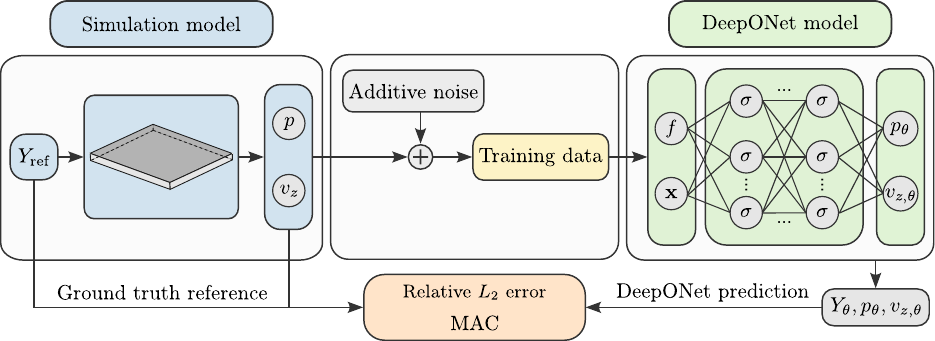}
		\caption{\label{fig:Method_scheme}{Schematic overview of the proposed methodology. A numerical simulation model generates reference acoustic field data using a prescribed surface admittance $Y_{\mathrm{ref}}$. Complex sound pressure $p$ and particle velocity in $z$-direction $v_z$ are extracted at discrete spatial locations and corrupted with additive noise to emulate measurement conditions. The resulting dataset, together with the input frequency $f$ and spatial coordinates $\mathbf{x}$, is used to train a DeepONet model that learns the operator mapping to the acoustic field ($p_{\theta}$, $v_{z,\theta}$) and the prediction of the surface admittance $Y_{\theta}$. Model accuracy is assessed by comparing the predicted admittance with the reference using the relative $L_2$-norm error and the MAC.}}
\end{figure*}

Fig.~\ref{fig:Method_scheme} provides a schematic overview of the proposed methodology.
A numerical simulation model is employed to generate reference acoustic field data based on a prescribed surface admittance $Y_{\mathrm{ref}}$. The formulation of this model is described in detail in Sec.~\ref{sec: Model description}. From the simulated sound field, complex-valued sound pressure $p$ and particle velocity in perpendicular direction to the sample's surface $v_z$ are extracted at discrete spatial locations. To emulate realistic measurement conditions, additive noise is superimposed on the simulated quantities, yielding the training dataset used for model training. The noisy pressure and particle velocity data, together with the corresponding input frequency $f$ and spatial measurement coordinates $\mathbf{x}$, are provided to the DeepONet, which is trained to learn the underlying operator mapping to the acoustic field quantities $p_\theta$ and $v_{z,\theta}$ and the surface admittance $Y_\theta$. The physics-informed DeepONet architecture utilized in this work is introduced in Sec.~\ref{sec:Physics-informed neural operators}. The resulting predictions are analyzed and discussed in Sec.~\ref{sec: Results and discussion}. Model performance is assessed by comparing the predicted admittance with the reference spectrum using the relative $L_2$-norm error and the modal assurance criterion (MAC). Finally, Sec.~\ref{sec:Conclusions} summarizes the findings and discusses the advantages and limitations of the proposed approach.

\section{\label{sec: Model description}Model description}
\begin{figure}[tbh]
	\centering
	\includegraphics[width=0.95\columnwidth]{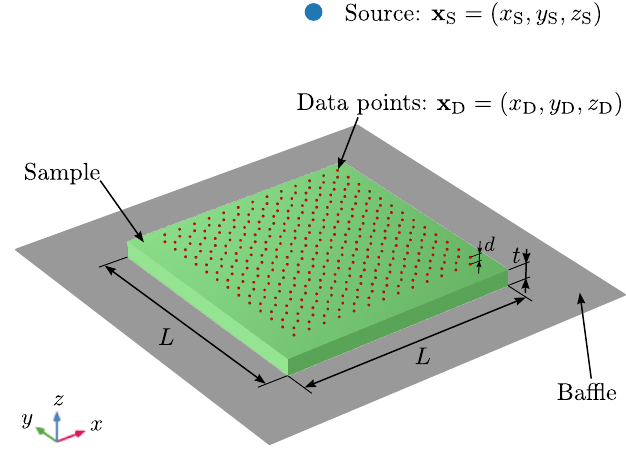}
	\caption{Schematic illustration of the model used to simulate a typical free-field measurement configuration. A monopole source is located at $\mathbf{x}_S$ (blue), while the sound-absorbing sample of dimensions $L \times L \times t$ (green) is mounted flush on a rigid baffle at $z=0$. Acoustic data are recorded at discrete measurement positions (red dots) arranged in two microphone arrays, which are separated by a distance $d$ along the $z$-direction.}
	\label{fig:Model_sketch}
\end{figure}

Synthetic measurement data are generated using a numerical boundary element method (BEM) model of a finite, planar porous material sample under semi-free-field conditions. This configuration is chosen to closely replicate typical free-field measurement scenarios. The square sample has lateral dimensions of $L=\SI{50}{cm}$ and a thickness of $t=\SI{3.5}{cm}$. A schematic representation of the model setup is shown in Fig.~\ref{fig:Model_sketch}.

Acoustic excitation is introduced by a monopole source with a pressure amplitude of $p_S=\SI{1}{\Pa}$, located at $\mathbf{x}_S = [0,\, 0,\,1]\,\mathrm{m}$. The surrounding medium is modeled as air with a density of $\rho_0=\SI{1.2}{\kg\per\cubic\m}$ and a speed of sound of $c=\SI{343}{\m\per\s}$, yielding a characteristic acoustic impedance \(Z_0 = \rho_0 c\). The material sample is placed flush on an idealized infinite rigid baffle located at $z = 0$. The sample is centered at the origin of the coordinate system. On the exposed surface of the sample (indicated in green in Fig.~\ref{fig:Model_sketch}), boundary conditions are specified as Robin conditions with a spatially uniform, complex-valued surface admittance \(\widetilde{Y}\). Throughout this work, the surface admittance is expressed in dimensionless form as \(Y = \widetilde{Y} Z_0\). To ensure physically realistic boundary conditions, the frequency-dependent reference admittance values $Y_{\mathrm{ref}}$ are obtained from impedance tube measurements of two different materials, as described in Sec.~\ref{subsec:Impedance tube measurements}. The simulation model is implemented using the commercial software COMSOL Multiphysics\textsuperscript{\tiny{\textregistered}} (COMSOL Inc., Stockholm, Sweden). The sample surface is discretized using triangular elements of quadratic order, with a mesh resolution corresponding to seven elements per wavelength at the highest frequency considered. Simulations are performed over a frequency range from $100$ to \SI{5000}{Hz} at discrete frequencies corresponding to the center frequencies of the \mbox{1/12th} octave bands. 

From the simulated dataset, the sound pressure $p$ and the particle velocity in the $z$-direction $v_{z}$ are evaluated at $N_D$ discrete observation points $\mathbf{x}_D$. In practical measurements, these quantities can be acquired simultaneously using a PU-probe. The observation points are arranged in a double-layer planar sensor array centered at the origin of the coordinate system. The two array layers are separated by $d = \SI{1.5}{cm}$ in the $z$-direction, while the lower layer is positioned at a distance of $h =\SI{0.5}{cm}$ above the sample surface. Within each layer, the sensors are placed on an equidistant xy-grid with a spacing of \SI{3.33}{cm}. Based on the parameter study results presented in Sec.~\ref{subsec:Parameter studies}, an array configuration with $13 \times 13$ sensors per layer is adopted. This leads to a total of $N_D = 338$ measurement points and an overall array aperture of $40$ $\times$ \SI{40}{cm}. The resulting sensor configuration is illustrated by red dots in Fig.~\ref{fig:Model_sketch}. To emulate realistic measurement condition and account for measurement noise, zero-mean Gaussian noise is added to the simulated data at each discrete frequency, with the noise level defined by a prescribed signal-to-noise ratio (SNR). The noise is applied independently to the sound pressure and particle velocity data, as well as to their respective real and imaginary components. Based on the results of the parameter study presented in Sec.~\ref{subsec:Parameter studies}, an SNR of \SI{40}{dB} is adopted throughout this work.

\subsection{\label{subsec:Impedance tube measurements} Impedance tube measurements}
\begin{figure}[htb]
	\centering
	\includegraphics[width=\columnwidth]{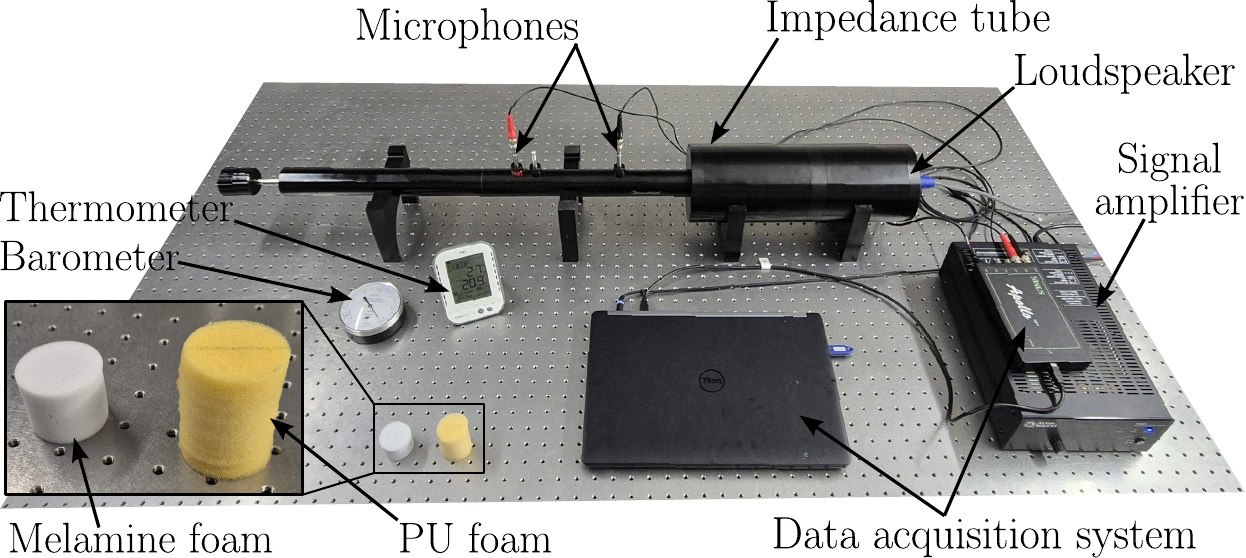}
	\caption{Measurement setup of the two-microphone impedance tube in accordance with ISO~10534-2~\cite{ISO105342.1998}. The insert in the lower left corner provides a magnified view of the investigated material samples, showing the melamine foam (left) and the PU foam (right).}  
	\label{fig:impedance_tube}
\end{figure}

Experimentally measured surface admittances are assigned as frequency-dependent, spatially uniform reference values $Y_{\mathrm{ref}}$ on the sample surface, as described in Sec.~\ref{sec: Model description}. The measurements are obtained using a standardized two-microphone impedance tube setup based on the transfer function method in accordance with ISO~10534-2~\cite{ISO105342.1998}. Fig.~\ref{fig:impedance_tube} shows a picture of the complete experimental setup. Measurements are conducted using an AED AcoustiTube impedance tube system (Gesellschaft für Akustikforschung Dresden mbH, Dresden, Germany) with an inner diameter of \SI{40}{mm}, providing a usable frequency range from $100$ to \SI{5000}{Hz}. The setup is equipped with two 1/4'' microphones of type M370 (Microtech Gefell GmbH, Gefell, Germany). Prior to the measurements, the microphones are calibrated using a pistonphone calibrator type 4228 (Brüel \& Kjær, Virum, Denmark). Signal acquisition is performed with an Apollo Light data acquisition system (SINUS Messtechnik GmbH, Leipzig, Germany), while the excitation signal is driven by a single-channel power amplifier PA601 (AtlasIED, Phoenix, USA). Throughout the experiments, environmental parameters including temperature (\SI{21.2}{\degreeCelsius}), relative humidity (\SI{32}{\%}), and barometric pressure (\SI{96.6}{kPa}) are continuously monitored and remained stable.
Two porous materials commonly used in noise control and room acoustic treatment are investigated in this study. A picture of the material samples is included in the lower left corner of Fig.~\ref{fig:impedance_tube}. Sample~1, displayed on the left, consists of melamine foam with a thickness of $\SI{35}{mm}$. Sample~2, shown on the right, is a polyurethane (PU) foam with a thickness of $\SI{55}{mm}$. Both samples are cut to a diameter of \SI{40}{mm} to ensure an accurate fit within the impedance tube. To account for potential variability associated with mounting conditions, each sample is measured three times, with the specimen being removed and reinstalled between successive measurements. The reported surface admittance curves represent the average of these repeated measurements, thereby providing a reliable reference for the subsequent analysis. The obtained reference surface admittance values are depicted as dash-dotted lines for $\Re(Y_{\mathrm{ref}})$ and dotted lines for $\Im(Y_{\mathrm{ref}})$ in Fig.~\ref{fig:Y_estimation}.

\section{\label{sec:Physics-informed neural operators}Physics-informed neural operators}
Neural operators constitute a class of machine learning methods that aim to approximate mappings between infinite-dimensional function spaces. Such mappings frequently appear in the mathematical description of physical systems, for instance when relating boundary conditions, spatially varying coefficients, or source terms of PDEs to their corresponding solution fields. In the present context, the nonlinear solution operator $\mathcal{G}$ maps a frequency-dependent input function $f \in \mathcal{F}$ to the associated acoustic solution field $s(\mathbf{x}) \in \mathcal{S}$ according to
\begin{equation}
 	\label{eq:no_goal}
 	\mathcal{G}: f \mapsto s(\mathbf{x}), \qquad \mathbf{x} \in \Omega ,
\end{equation}
where $\mathbf{x} = [x, y, z]^{T}$ denotes the spatial coordinate and $\Omega$ represents the computational domain. The solution vector $\mathbf{s}(\mathbf{x})$ of the acoustic field comprises both the sound pressure field $p(\mathbf{x})$ and the particle velocity field in $z$-direction $v_z(\mathbf{x})$.

The objective of operator learning is to construct a parameterized surrogate operator $\mathcal{G}_\theta : \mathcal{F} \rightarrow \mathcal{S}$ that approximates the unknown operator $\mathcal{G}$. The surrogate is represented by a neural network with trainable parameters $\theta \in \mathbb{R}^{l}$ and yields the approximation
\begin{equation}
 	\label{eq:no_approx}
 	\mathcal{G}_\theta(f, \mathbf{x}) = \mathbf{s}_\theta(f, \mathbf{x}) \approx \mathcal{G}(f, \mathbf{x}),
 	\qquad \forall f \in \mathcal{F}, \; \mathbf{x} \in \Omega .
\end{equation}
Here, $l \in \mathbb{N}$ denotes the number of trainable parameters and $\mathbf{s}_\theta(f, \mathbf{x})$ represents the neural operator prediction of the acoustic solution field. The parameters $\theta$ are determined during training by minimizing a supervised loss function $\mathcal{L}(\theta)$ that measures the discrepancy between the neural operator predictions and corresponding reference data on a finite training dataset. This leads to the optimization problem
\begin{equation}
 	\label{eq:no_optimization}
 	\theta^{*} = \arg\min_{\theta} \mathcal{L}\big(\mathcal{G}_\theta(f, \mathbf{x}), \mathcal{G}(f, \mathbf{x})\big),
\end{equation}
where $\theta^{*}$ denotes the set of optimized weights and biases of the network after convergence of the training procedure.
 
In practical applications, neural operators provide a flexible and powerful framework for approximating solution operators associated with parameterized PDEs. One key advantage of this class of models is their resolution invariance, meaning the learned operator can be trained and evaluated on grids of different spatial discretization without modifying the network architecture~\cite{Azizzadenesheli.2024}. Furthermore, in contrast to PINNs, neural operators do not require retraining when new input functions are considered. Once the model has been trained, it can be directly evaluated for previously unseen inputs, which enables significant benefits in settings that involve repeated queries or near real-time predictions~\cite{Rosofsky.2023}. Another important property is that neural operator models are continuously differentiable with respect to their inputs. This characteristic facilitates gradient-based sensitivity analyses and allows derived physical quantities to be computed efficiently using automatic differentiation, an aspect that is exploited in the present work. Among the various neural operator architectures proposed in recent years, the DeepONet introduced by Lu et al.~\cite{Lu.2021} is employed in the present work.

\subsection{\label{subsec:Deep Operator Network} Deep operator network (DeepONet)}
\begin{figure*}[tbh]
	\begin{center}
		\includegraphics[width = 0.9\textwidth]{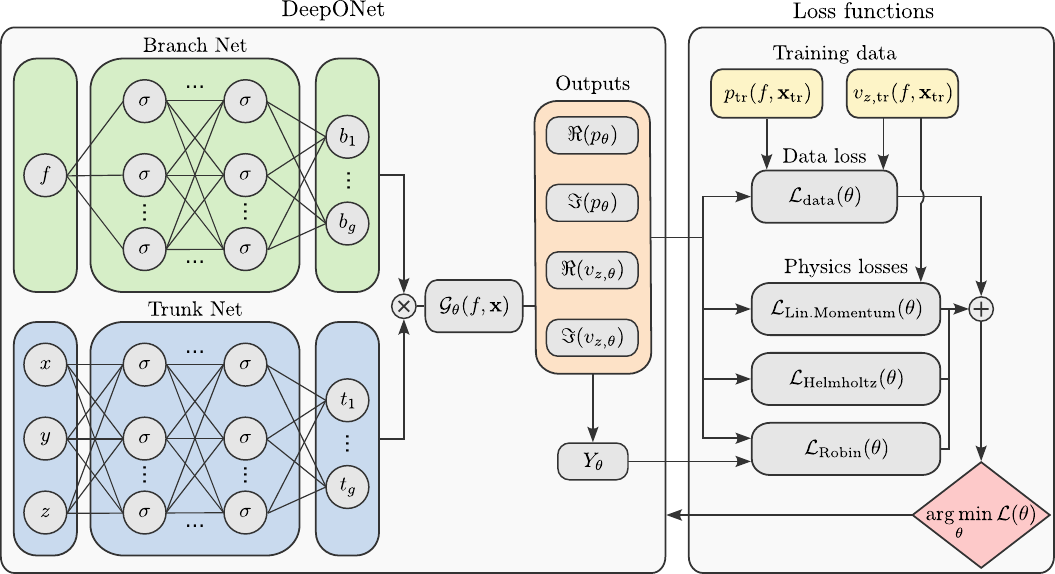}
		\caption{\label{fig:DeepONet_scheme}{Schematic overview of the physics-informed DeepONet architecture and training procedure. The neural operator consists of a branch network and a trunk network that together map the input frequency and spatial coordinates to the predicted acoustic field quantities (left). The resulting network outputs (center) are used to evaluate the different loss components that guide the training process, including the supervised data loss and the individual physics-informed residual terms (right). The overall loss function \( \mathcal{L}(\theta) \) is defined as the weighted sum of these individual loss contributions and is minimized during training to obtain a neural operator that simultaneously fits the training data and satisfies the governing physical equations.
		}}
	\end{center}
\end{figure*}

DeepONets, whose theoretical foundation is provided by the universal approximation theorem for nonlinear operators~\cite{Chen.1995}, provide a practical architecture for learning nonlinear operators from data~\cite{Rosofsky.2023}. The central idea is to separate the operator learning task into two complementary neural sub-networks. The branch network (Fig.~\ref{fig:DeepONet_scheme}, green) encodes the input frequency function by processing its discretized representation 
\[
f^{(i)} = \left[ f^{(1)}, f^{(2)}, \ldots, f^{(N_f)} \right]^T \in \mathbb{R}^{N_f},
\]
and maps it to a latent feature vector $b_k = [b_1, b_2, \ldots, b_{g}]^T \in \mathbb{R}^{g}$. In parallel, the trunk network (Fig.~\ref{fig:DeepONet_scheme}, blue) evaluates spatial information by taking a coordinate $\mathbf{x} \in \Omega \subset \mathbb{R}^{N_{\mathrm{tr}}}$ as input and returns a corresponding feature vector $t_k = [t_1, t_2, \ldots, t_{g}]^T \in \mathbb{R}^{g}$. Here, $N_f$ denotes the number of discrete frequency samples considered in the dataset, while $N_{\mathrm{tr}}$ represents the total number of spatial training points. The predicted operator output at a given input frequency $f$ and spatial location $\mathbf{x}$ is obtained by combining the two latent representations through an inner product. This interaction couples the functional information captured by the branch network with the spatial basis encoded by the trunk network, yielding the DeepONet approximation
\begin{equation}
	\label{eq:deeponet}
	\mathcal{G}_{\theta}(f, \mathbf{x})
	=
	\sum_{k=1}^{g}
	\underbrace{b_k\!\left(f^{(1)}, f^{(2)}, \ldots, f^{(N_f)}\right)}_{\text{branch}}
	\cdot
	\underbrace{t_k(\mathbf{x})}_{\text{trunk}} \, .
\end{equation}
This formulation of the operator evaluation can be conceptually interpreted as a data-driven spectral expansion, where the trunk network provides the spatial basis functions and the branch network determines the expansion coefficients conditioned on the input function~\cite{Ray.2024}. By leveraging this disentanglement of spatial and frequency dependencies, the DeepONet architecture decomposes the overall operator learning task into two coupled but simpler subproblems. This separation of roles reduces the complexity of the learning problem and enables the model to generalize to arbitrary spatial evaluation points~\cite{Lu.2021}.

The DeepONet is trained using a dataset composed of frequency-dependent input functions, spatial sampling locations, and the corresponding acoustic field quantities. For each discrete frequency $f^{(i)}$ and spatial training point $\mathbf{x}_{\mathrm{tr}}^{(j)}$, the dataset contains the complex-valued sound pressure $p_{\mathrm{tr}}\!\left(f^{(i)}, \mathbf{x}_{\mathrm{tr}}^{(j)}\right)$ as well as the particle velocity component in $z$-direction $v_{z, \mathrm{tr}}\!\left(f^{(i)}, \mathbf{x}_{\mathrm{tr}}^{(j)}\right)$. The resulting training dataset can be written as
\begin{equation}
	\label{eq:training_data}
	\mathcal{D}_{\mathrm{tr}} =
	\left\{
	f^{(i)},\,
	\mathbf{x}_{\mathrm{tr}}^{(j)},\,
	p_{\mathrm{tr}}\!\left(f^{(i)}, \mathbf{x}_{\mathrm{tr}}^{(j)}\right), 
	v_{z, \mathrm{tr}}\!\left(f^{(i)}, \mathbf{x}_{\mathrm{tr}}^{(j)}\right)
	\right\},
\end{equation}
where $i = 1,\ldots,N_f$ and $j = 1,\ldots,N_{\mathrm{tr}}$. 

\subsection{\label{subsec:Physics-informed loss function} Physics-informed loss function}
Conventional data-driven DeepONets rely solely on paired input - output data for training. As a consequence, their predictive accuracy is strongly influenced by the size and diversity of the available dataset, and there is generally no guarantee that the learned operator satisfies the underlying physical laws. Physics-informed DeepONets address this limitation by embedding the governing physical equations directly into the training objective~\cite{Wang.2021b}. In practice, residuals of the relevant physical equations are incorporated into the loss function, enabling the network to learn simultaneously from data and from physical constraints. This additional physics-based regularization guides the optimization process, leading to improved physical consistency of the predictions and enhanced generalization capabilities of the trained model~\cite{Li.2024}.

As visually illustrated in Fig.~\ref{fig:DeepONet_scheme}, the total loss function \( \mathcal{L}(\theta) \) is defined as a weighted sum of data-driven and physics-based loss terms,
\begin{equation}
	\begin{aligned}
		\mathcal{L}(\theta) =
		\lambda_p \mathcal{L}_p(\theta)
		+
		\lambda_{v_z} \mathcal{L}_{v_z}(\theta)
		+
		\lambda_{\mathrm{LM}}\,\mathcal{L}_{\mathrm{LM}}(\theta)
		\\
		+
		\lambda_{\mathrm{Helm}}\,\mathcal{L}_{\mathrm{Helm}}(\theta)
		+
		\lambda_{\mathrm{R}}\,\mathcal{L}_{\mathrm{R}}(\theta),
	\end{aligned}
\end{equation}
where \( \mathcal{L}_p(\theta) \) and \( \mathcal{L}_{v_z}(\theta) \) represent the supervised data losses for the sound pressure and the particle velocity in \(z\)-direction, respectively. The physics-informed loss terms promote consistency with the governing equations by penalizing their respective residuals. In particular, \( \mathcal{L}_{\mathrm{LM}}(\theta) \) enforces the linearized momentum equation, \( \mathcal{L}_{\mathrm{Helm}}(\theta) \) ensures compliance with the Helmholtz equation in the acoustic domain, and \( \mathcal{L}_{\mathrm{R}}(\theta) \) imposes the Robin boundary condition at the sample's surface. The weighting factors \( \lambda_p \), \( \lambda_{v_z} \), \( \lambda_{\mathrm{LM}} \), \( \lambda_{\mathrm{Helm}} \), and \( \lambda_{\mathrm{R}} \) control the relative contribution of the individual loss terms and balance the trade-off between data fidelity and physical consistency during training.

\paragraph*{Data losses:}
The supervised data losses are formulated as mean squared error (MSE) between the neural operator predictions and the corresponding values from the training dataset. For the sound pressure, the loss is defined as
\begin{equation}
	\mathcal{L}_p(\theta)
	=
	\frac{1}{N_d}
	\sum_{n=1}^{N_d}
	\Big|
	p_{\theta}(f_n,\mathbf{x}_n)
	-
	p_{\mathrm{tr}}(f_n,\mathbf{x}_n)
	\Big|^2,
\end{equation}
while the particle-velocity loss is given as
\begin{equation}
	\mathcal{L}_{v_z}(\theta)
	=
	\frac{1}{N_d}
	\sum_{n=1}^{N_d}
	\Big|
	v_{z,\theta}(f_n,\mathbf{x}_n)
	-
	v_{z,\mathrm{tr}}(f_n,\mathbf{x}_n)
	\Big|^2 .
\end{equation}
Here, \(N_d = N_f N_{\mathrm{tr}}\) denotes the total number of supervised training samples, obtained from all combinations of the \(N_f\) frequencies and the \(N_{\mathrm{tr}}\) spatial training locations.

\paragraph*{Linearized momentum loss:}
To enforce consistency between the predicted pressure and the measured particle velocity field, a linearized momentum loss is introduced. Under the harmonic time dependence \(e^{\mathrm{i}\omega t}\) (with \(\mathrm{i}^2=-1\)), the linearized momentum equation in frequency domain relates the pressure derivative in $z$-direction to the particle velocity according to
\begin{equation}
	\label{eq:Lin. Momentum}
	\frac{\partial p(f,\mathbf{x})}{\partial z}
	=
	-\,\mathrm{i} \omega \rho_0\, v_z(f,\mathbf{x}),
\end{equation}
where \( \omega = 2\pi f \). The corresponding loss term penalizes deviations from this physical relation and is defined as
\begin{equation}
	\mathcal{L}_{\mathrm{LM}}(\theta)
	=
	\frac{1}{N_d}
	\sum_{n=1}^{N_d}
	\left|
	\frac{\partial p_{\theta}(f_n,\mathbf{x}_n)}{\partial z}
	+
	i \omega_n \rho_0\, v_{z,\mathrm{tr}}(f_n,\mathbf{x}_n)
	\right|^2 .
\end{equation}
Importantly, the linearized momentum loss is evaluated at the same spatial locations and frequencies as the supervised training data. Consequently, the training points simultaneously serve as data samples and physics collocation points. In this way, the measurements constrain not only the pressure field but also its spatial gradient in the \(z\)-direction through the measured particle velocity, thereby improving the physical consistency and stability of the learned operator.

\paragraph*{Helmholtz loss:}
The acoustic field above the sample is governed by the homogeneous Helmholtz equation. In this work, a mixed formulation of the Helmholtz equation is employed, in which the linearized momentum Eq.~(\ref{eq:Lin. Momentum}) is substituted into the Helmholtz equation to express the pressure derivative in the \(z\)-direction in terms of particle velocity \(v_z\). This mixed formulation reduces the derivative order in \(z\)-direction that must be evaluated with respect to the neural operator outputs. As higher-order gradients are known to be prone to numerical instabilities and exploding gradients when computed through automatic differentiation, the computation of the physics-informed residuals becomes numerically more stable and accurate~\cite{Wang.2021}. The resulting governing equation reads
\begin{equation}
	\frac{\partial^2 p(f,\mathbf{x})}{\partial x^2}
	+
	\frac{\partial^2 p(f,\mathbf{x})}{\partial y^2}
	-
	i\omega\rho_0 \frac{\partial v_z(f,\mathbf{x})}{\partial z}
	+
	k^2 p(f,\mathbf{x})
	=
	0,
\end{equation}
where \( k = \omega / c_0 \) denotes the wave number in the acoustic domain. The corresponding collocation-based Helmholtz loss penalizes violations of this governing equation and is defined as
\begin{equation}
	\begin{aligned}
		\mathcal{L}_{\mathrm{Helm}}(\theta)
		=
		\frac{1}{N_c}
		\sum_{m=1}^{N_c}
		\Big|
		\frac{\partial^2 p_{\theta}(f_m,\mathbf{x}_m)}{\partial x^2}
		+
		\frac{\partial^2 p_{\theta}(f_m,\mathbf{x}_m)}{\partial y^2}\\
		-
		\mathrm{i}\omega_m\rho_0
		\frac{\partial v_{z,\theta}(f_m,\mathbf{x}_m)}{\partial z}
		+
		k_m^2 p_{\theta}(f_m,\mathbf{x}_m)
		\Big|^2 ,
	\end{aligned}
\end{equation}
where \(N_c\) denotes the number of collocation points. The Helmholtz residual is evaluated at collocation points within the acoustic domain that are randomly sampled during training from a predefined spatial grid. Specifically, the collocation points are drawn from the spatial region \(x \in [-L/2,\,L/2]\), \(y \in [-L/2,\,L/2]\), and \(t < z < (t+h+d)\), i.e., the domain between the absorber surface and the highest measurement plane within the lateral extent of the measurement array. Notably, the Helmholtz loss is evaluated using only the neural operator predictions, such that the governing wave equation is enforced as a self-consistency constraint on the predicted acoustic fields.

\paragraph*{Robin boundary loss:}
Finally, the unknown normalized surface admittance \(Y_{\theta}(f)\) is learned jointly with the DeepONet by enforcing a spatially uniform Robin boundary condition at the sample surface \(\Gamma_Y\) at $z=t$. The admittance boundary condition can be written as
\begin{equation}
	\frac{\partial p(f,\mathbf{x})}{\partial n} + \mathrm{i} k Y(f)\,p(f,\mathbf{x}) = 0 	\qquad \text{on } \Gamma_Y .
\end{equation}
The acoustic domain is located above the sample surface such that the outward normal of the fluid domain points in the negative \(z\)-direction. Using \(\partial_n p = -\partial_z p\) together with the linearized momentum relation in Eq.~(\ref{eq:Lin. Momentum}) yields
\begin{equation}
	\mathrm{i}\omega\rho_0\,v_z(f,\mathbf{x})
	+
	\mathrm{i}kY(f)\,p(f,\mathbf{x})
	=
	0
	\qquad \text{on } \Gamma_Y .
\end{equation}
The corresponding Robin boundary loss penalizes violations of this admittance relation and is defined as
\begin{equation}
	\mathcal{L}_{\mathrm{R}}(\theta)
	=
	\frac{1}{N_{bc}}
	\sum_{l=1}^{N_{bc}}
	\Big|
	\mathrm{i}\omega_l\rho_0\,v_{z,\theta}(f_l,\mathbf{x}_l)
	+
	\mathrm{i}k_l Y_{\theta}(f_l)\,p_{\theta}(f_l,\mathbf{x}_l)
	\Big|^2 ,
\end{equation}
where \(N_{bc}\) denotes the number of boundary collocation points. The boundary residual is evaluated at collocation points randomly drawn during training from a predefined spatial grid on the sample surface, specifically within the spatial region \(x\in[-L/2,L/2]\), \(y\in[-L/2,L/2]\), and \(z=t\), corresponding to the absorber surface within the lateral extent of the measurement array.

Importantly, the surface admittance \(Y_{\theta}(f)\) is treated as a trainable parameter and optimized simultaneously with the neural operator parameters while training. During gradient-based optimization, the admittance parameters are updated together with the network weights such that the predicted pressure and particle velocity jointly satisfy the admittance boundary condition at all sampled boundary collocation points. In this formulation, the admittance estimation is embedded directly into the physics-informed training process and therefore acts as a strong regularizer for the underlying physics-constrained inverse problem. Rather than computing the admittance in a post-processing step from point-wise ratios of particle velocity and pressure, the parameter \(Y_{\theta}(f)\) is inferred as the value that globally minimizes the boundary residual on the sample surface while remaining consistent with the governing equations. This global optimization substantially improves the robustness and stability of the inferred admittance compared to local point-wise estimates, which are typically sensitive to measurement noise and small pressure amplitudes.

\subsection{\label{subsec:DeepONet architecture and training} DeepONet architecture and training}
The physics-informed DeepONet framework and the associated training workflow are schematically illustrated in Fig.~\ref{fig:DeepONet_scheme}. The left part of Fig.~\ref{fig:DeepONet_scheme} depicts the DeepONet architecture, which consists of a branch network and a trunk network. The network architecture and associated hyperparameters are selected based on an extensive grid-search hyperparameter optimization. Both subnetworks are implemented as residual multilayer perceptrons comprising three hidden layers with 96 neurons per layer. Sinusoidal activation functions employed throughout the networks using the SIREN formulation~\cite{Sitzmann.2020}, enabling the representation of highly oscillatory functions such as acoustic wave fields. To ensure stable training of sinusoidal networks, the weights are initialized using the SIREN initialization scheme~\cite{Sitzmann.2020}. In addition, residual skip connections are introduced between hidden layers to improve gradient propagation and stabilize the optimization of deep sinusoidal networks. The output dimension of both the branch and trunk networks is set to \(g = 128\). Their outputs are combined through an element-wise product and subsequently mapped by a linear output layer to the predicted acoustic quantities. Specifically, the network predicts the real and imaginary parts of the acoustic pressure \(p_{\theta}\) and the normal-direction particle velocity \(v_{z,\theta}\), as illustrated in Fig.~\ref{fig:DeepONet_scheme}.

The relative contributions of the individual loss components are governed by weighting coefficients, which are determined through an extensive grid-search-based hyperparameter optimization. The final values utilized in this work are \(\lambda_p = 0.108\), \(\lambda_{v_z} = 1.0\), \(\lambda_{\mathrm{LM}} = 2.80 \times 10^{-3}\), \(\lambda_{\mathrm{Helm}} = 4.23 \times 10^{-8}\), and \(\lambda_{\mathrm{R}} = 2.896\). To enhance training stability, the physics-informed loss components are introduced progressively using cosine ramp functions. This strategy ensures that the optimization initially prioritizes the fitting of measurement data before gradually enforcing the governing equations and boundary conditions. Specifically, the linearized momentum loss is activated between $20\%$ and $40\%$ of the total training epochs, followed by the Helmholtz loss between $30\%$ and $60\%$, and finally the Robin boundary condition loss between $55\%$ and $85\%$.

To ensure numerical stability and balanced gradient contributions across frequencies and loss components, both the training data and the physics-informed residuals are normalized. The real and imaginary parts of the sound pressure \(p\) and the particle velocity \(v_z\) are normalized separately for each frequency using a frequency-wise \(z\)-score normalization computed from the training data. Consequently, the neural operator predicts normalized field quantities during training, while the physical variables are recovered through inverse scaling when evaluating the physics-informed residuals. Further, the residuals of the Helmholtz equation, the linearized momentum equation, and the Robin boundary condition are normalized using frequency-dependent reference powers derived from the measured field amplitudes. This strategy prevents frequencies with large field amplitudes from dominating the optimization and ensures that the different loss components remain on comparable numerical scales during training. In addition, the spatial coordinates are linearly normalized to the interval \([-1,1]\), while the frequency input is log-transformed and standardized using a \(z\)-score normalization based on the training data.

The model parameters are optimized using the AdamW optimizer~\cite{Loshchilov.2019} with an initial learning rate of \(\eta_{\mathrm{init}} = 10^{-3}\) and a weight decay of \(10^{-5}\). A cosine annealing schedule is applied to gradually decrease the learning rate to \(\eta_{\mathrm{min}} = 2 \times 10^{-4}\) during training. The surface admittance parameter $Y_{\theta}$ is optimized with a constant learning rate of \(10^{-4}\). Training is performed for \(50{,}000\) epochs using mini-batch optimization with a batch size of \(8192\) for the supervised data losses. The physics-informed losses are evaluated at randomly sampled collocation points within a specified spatial grid, with \(4096\) interior points for the Helmholtz residual and \(4096\) boundary points for the Robin boundary condition. To reduce computational cost, the Helmholtz and linearized momentum losses are evaluated only at every fourth training step. Gradient clipping with a threshold of \(1.0\) is applied to prevent unstable parameter updates caused by exploding gradients.

The training dataset consists of collocated samples of the real and imaginary parts of the acoustic pressure \(p\) and the normal-direction particle velocity \(v_z\) obtained from \(90\%\) of the available measurement points \(N_D = 338\), corresponding to \(N_{\mathrm{tr}} = 304\) spatial training locations. The data cover the frequency range from $100$ to \SI{5000}{Hz} at discrete frequencies corresponding to the center frequencies of the \mbox{1/12th} octave bands. Model performance is monitored on a validation dataset generated by randomly selecting \(10\%\) of the available measurement points without replacement, ensuring that the validation samples remain disjoint from the training set. It is worth noting that the present study focuses on spatial generalization, whereas frequency generalization is not explicitly targeted. This choice is motivated by the fact that the dominant experimental effort arises from spatial scanning of the measurement grid, whereas increasing the spectral resolution does not impose a significant additional cost for modern data acquisition systems. The DeepONet model is implemented in PyTorch~\cite{Paszke.2019} and optimized for execution on an NVIDIA GeForce RTX 5080 graphics processing unit (GPU). Training the network for \(50{,}000\) epochs requires approximately \(35\,\mathrm{min}\) of wall-clock time.

\begin{figure}[htb]
	\centering
	\includegraphics[width=\columnwidth]{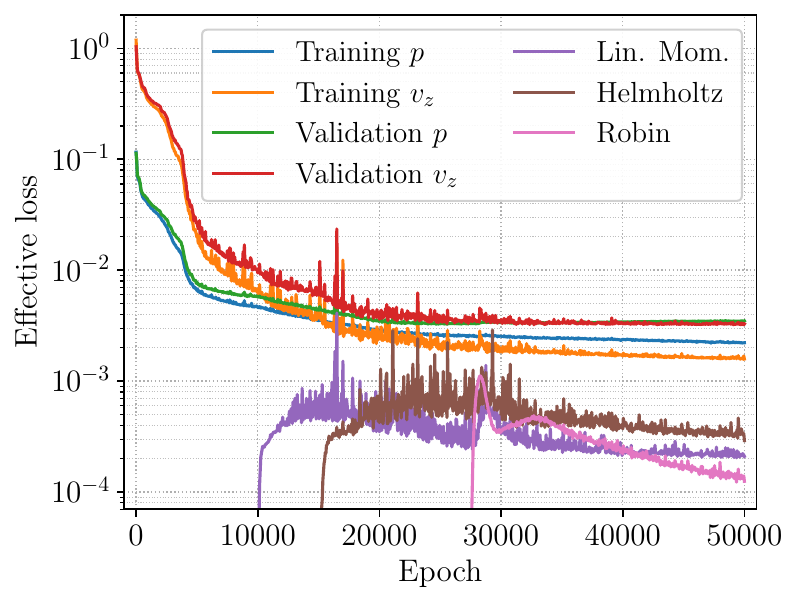}
	\caption{Evolution of the effective training and validation losses (\(\lambda \cdot \mathcal{L}\)) during DeepONet training. The supervised data losses for pressure and particle velocity decrease rapidly and converge smoothly, while the physics-informed residual losses appear progressively as their weighting is activated during training. The curves demonstrate stable joint optimization of data-driven and physics-informed objectives.}
	\label{fig:Effective_loss}
\end{figure}
Fig.~\ref{fig:Effective_loss} shows the evolution of the effective loss (\(\lambda \cdot \mathcal{L}\)) components during training. The supervised data losses for the acoustic pressure \(p\) and particle velocity \(v_z\) decrease rapidly in the initial training phase and converge to low values, while the corresponding validation losses closely follow the training curves, indicating good generalization without signs of overfitting. The physics-informed loss terms initially increase upon activation and subsequently decrease as the network adapts to satisfy the additional physical constraints. Notably, the activation of the Helmholtz and linearized momentum losses initially leads to transient peaks, indicating the increased difficulty of enforcing the governing equations, before both terms stabilize and converge to lower magnitudes. Overall, the plot demonstrates well-balanced training, confirming that the network successfully learns a solution that simultaneously fits the measurement data and satisfies the underlying physical equations.

\section{\label{sec: Results and discussion}Results and discussion}
Upon completion of training, the physics-informed DeepONet provides continuous predictions of the acoustic field quantities at arbitrary spatial locations and frequencies, while simultaneously inferring a consistent global surface admittance spectrum of the absorber. For each of the two investigated material samples, a separate neural network is trained using an identical architecture and the same set of hyperparameters. Consequently, the proposed approach does not target generalization across different material types. Instead, each model is specifically trained to characterize the acoustic properties of the respective material under investigation. As the present study focuses exclusively on material characterization, all evaluations of the trained network are restricted to the specimen surface.
\begin{figure}[htb]
	\centering
	\includegraphics[width=0.95\columnwidth]{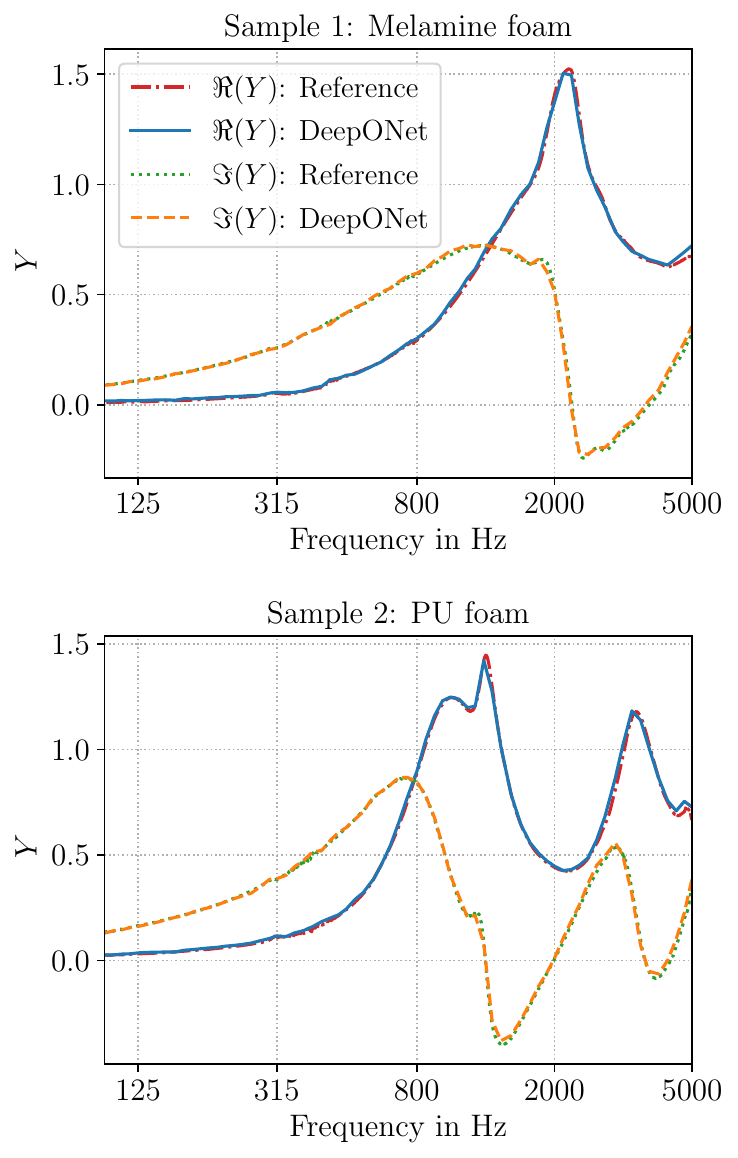}
	\caption{Comparison of the frequency-dependent real and imaginary parts of the acoustic surface admittance $Y$ for two porous materials: melamine foam (top) and PU foam (bottom). The DeepONet predictions are shown alongside the corresponding reference spectrum over the frequency range from $100$ to \SI{5000}{\Hz}.}
	\label{fig:Y_estimation}
\end{figure}

Fig.~\ref{fig:Y_estimation} shows the frequency-dependent real and imaginary parts of the surface admittance $Y$ for two representative porous materials, melamine foam (Sample~1, top) and PU foam (Sample~2, bottom). The DeepONet predictions are compared with the corresponding references over the frequency range from $100$ to \SI{5000}{\Hz}, covering more than $5.5$ octaves. The reference curves exhibit distinct acoustic characteristics of the two materials, which can be partly attributed to differences in their sample thicknesses. Overall, the model accurately reproduces the characteristic behavior of both $\Re(Y)$ and $\Im(Y)$ across the entire frequency range, including the pronounced peaks and frequency-dependent variations of both components. This demonstrates that both dissipative and reactive material properties are reliably captured.

The predictive performance of the DeepONet is assessed not only in terms of the estimated surface admittance, but also with respect to the sound pressure and particle velocity on the absorber surface. Two quantitative error metrics are employed, namely the MAC and the relative $L_2$-norm error of the acoustic quantities. The latter is defined as
\begin{align}
	L_2
	= 
	\frac{\sqrt{\sum_{i=1}^{N} \left| q_i^{\mathrm{pred}} - q_i^{\mathrm{ref}} \right|^2}}
	{\sqrt{\sum_{i=1}^{N} \left| q_i^{\mathrm{ref}} \right|^2}},
	\label{eq:L2_error}
\end{align}
where $q^{\mathrm{pred}}$ and $q^{\mathrm{ref}}$ denote the complex-valued predicted and reference quantities for the surface admittance $Y$, the sound pressure $p$, and the normal particle velocity $v_z$. For $p$ and $v_z$, the evaluation is performed at $3376$ boundary points, spanning the same $xy$-aperture as the measurement array and coinciding with the mesh nodes of the BEM model.
\begin{figure*}[tbh]
	\centering
	\includegraphics[width=\textwidth]{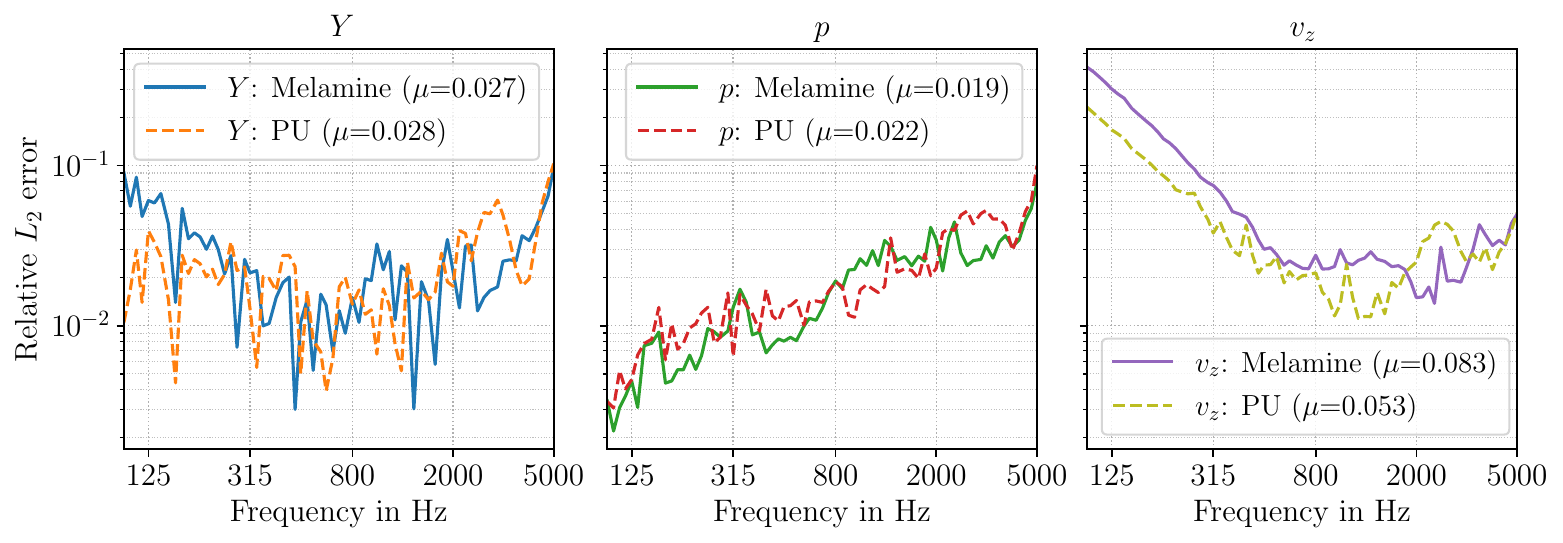}
	\caption{Relative $L_2$-norm error of the DeepONet predictions for the surface admittance $Y$ (left), the sound pressure $p$ (middle), and the normal particle velocity $v_z$ (right) for melamine foam and PU foam as a function of frequency from $100$ to \SI{5000}{\Hz}. The mean error values $\mu$ are reported in the legends.}
	\label{fig:Errors_Y_p_vz}
\end{figure*}

Fig.~\ref{fig:Errors_Y_p_vz} presents the relative $L_2$-norm error of the DeepONet predictions for the surface admittance $Y$, the sound pressure $p$, and the normal particle velocity $v_z$ for melamine and PU foam over the considered frequency range. The model achieves low errors for the admittance, with mean values of $0.027$ (melamine) and $0.028$ (PU), indicating an accurate reconstruction of the reference spectrum. The estimation is most accurate in the mid-frequency range, where the acoustic wavelength $\lambda$ is on the order of the characteristic specimen dimension $L$. In the low-frequency regime ($\lambda \gg L$), the sample behaves as an acoustically small object, and the influence of the admittance boundary condition on the pressure and particle velocity fields is weak. In addition, the absorptive behavior of porous materials is typically low in this range, further reducing sensitivity to boundary properties. Consequently, spatial gradients of the acoustic field are small and the measurements contain limited information, resulting in a poorly conditioned inverse problem. Additionally, the finite extent of the sample introduces deviations from idealized assumptions, such as edge diffraction effects, which further limit identifiability. The sound pressure field is similarly predicted with consistently low errors across frequencies, demonstrating robust generalization of the learned acoustic field. With increasing frequency, $\lambda$ decreases, leading to more complex interference patterns and stronger spatial variations in the pressure field, which in turn result in a gradual increase in the prediction error. 

In contrast, the particle velocity exhibits higher error levels, particularly at low frequencies. This behavior can be attributed to the increased sensitivity of particle velocity estimation to small pressure gradients and weak velocity amplitudes in this frequency regime, resulting in higher relative errors. Although $v_z$ is directly learned from data, it remains implicitly coupled to spatial pressure gradients through the governing equations enforced in the physics losses. As a result, inconsistencies between the predicted pressure and velocity fields are penalized during training, making the estimation of $v_z$ more susceptible to measurement noise and modeling inaccuracies. A comparison between the two materials further reveals that the particle velocity for melamine foam is less accurately predicted, especially in the low-frequency range. This can be attributed to its smaller sample thickness, which leads to reduced absorption and consequently lower particle velocity amplitudes on the surface. The same trend can also be observed in the surface admittance estimation results.

Across all quantities, the error generally increases at higher frequencies ($\lambda \ll L$), reflecting the increasingly complex spatial structure of the acoustic field, more intricate interference patterns, and the associated difficulty of resolving fine-scale variations. Although sensitivity to boundary properties improves in this regime, this advantage is partially offset by the finite spatial sampling of the measurement array, which may lead to undersampling of high-frequency content. In addition, more pronounced wave phenomena, such as localized effects and increase model approximation errors, further complicate the inverse problem. Nevertheless, the consistently low error levels indicate that the proposed approach remains robust and accurate across a wide frequency range.

To quantify the spatial agreement between the predicted and reference acoustic fields on the absorber surface, the MAC is employed. It is defined as
\begin{align}
	\mathrm{MAC} \;=\;
	\frac{\left| \boldsymbol{q}_{\mathrm{ref}}^{H} \, \boldsymbol{q}_{\mathrm{pred}} \right|^{2}}
	{\left( \boldsymbol{q}_{\mathrm{ref}}^{H} \boldsymbol{q}_{\mathrm{ref}} \right)
		\left( \boldsymbol{q}_{\mathrm{pred}}^{H} \boldsymbol{q}_{\mathrm{pred}} \right)},
	\label{eq:MAC}
\end{align}
where the superscript $H$ denotes the Hermitian. The MAC provides a normalized measure of spatial similarity for complex-valued field quantities. Its values range from 0, indicating no correlation, to 1, corresponding to perfect agreement. As such, it serves as a robust metric for evaluating the extent to which the predicted fields reproduce the spatial structure of the reference. For the surface admittance, the point-wise ratio between $v_z$ and $p$ is analyzed to assess the degree to which the spatial uniformity implied by the Robin boundary condition is satisfied.
\begin{figure*}[t]
	\centering
	\includegraphics[width=\textwidth]{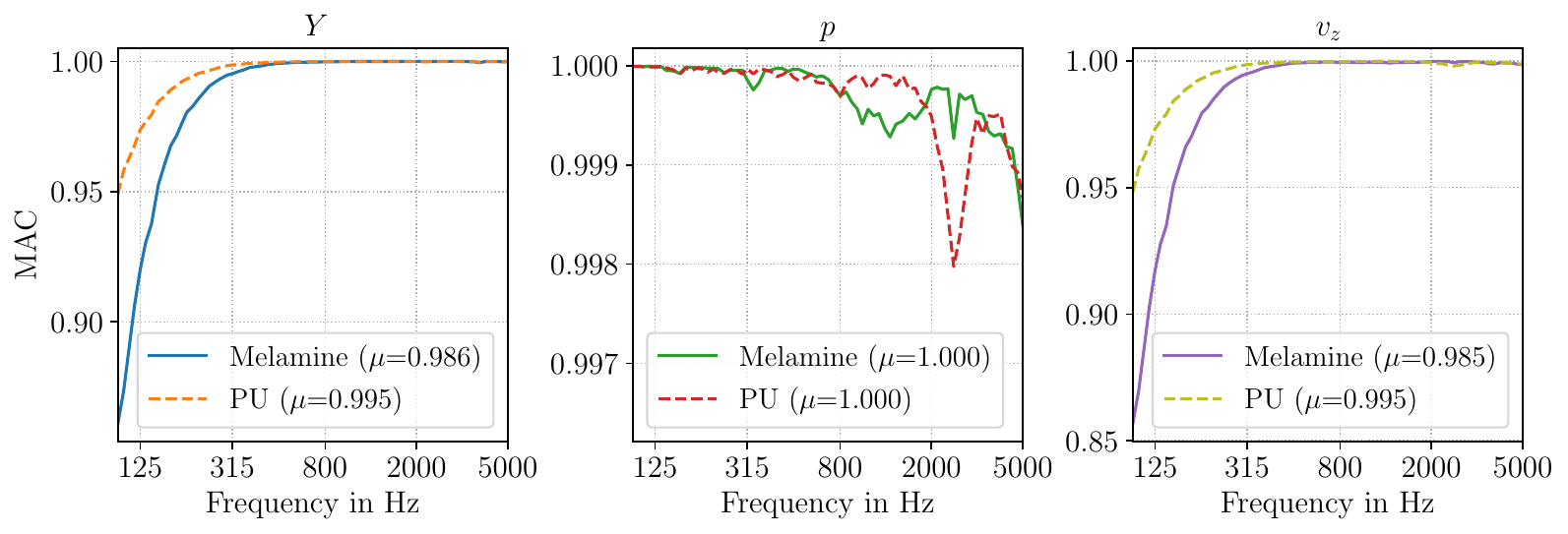}
	\caption{MAC between the DeepONet predictions and the reference for the surface admittance $Y$, the sound pressure $p$, and the normal particle velocity $v_z$ on the absorber surface for melamine foam and PU foam as a function of frequency from $100$ to \SI{5000}{\Hz}. The mean MAC values are indicated in the legends.}
	\label{fig:MAC_Y_p_vz}
\end{figure*}

Fig.~\ref{fig:MAC_Y_p_vz} presents the MAC between the DeepONet predictions and the numerically generated reference solutions for the surface admittance $Y$, the sound pressure $p$, and the normal particle velocity $v_z$ on the absorber surface for melamine and PU foam. Overall, the MAC values remain consistently close to unity across the considered frequency range, indicating excellent agreement in both amplitude and phase between the predicted and reference fields. For the sound pressure field, MAC values near unity are achieved, demonstrating that the spatial structure of the pressure field is reproduced with high fidelity. The particle velocity exhibits slightly lower MAC values and shows a degradation at low frequencies, where reduced information content and weaker spatial gradients limit identifiability. The surface admittance likewise shows strong agreement, with mean values of $0.986$ (melamine) and $0.995$ (PU), and a noticeable improvement from low to mid frequencies. The reduced MAC values at low frequencies are consistent with the elevated $L_2$ errors observed in this regime, as discussed previously. Moreover, differences between the two samples are also reflected in the MAC trends, particularly at low frequencies. Overall, the results confirm that the proposed approach accurately captures the spatial characteristics of all considered acoustic quantities across a wide frequency range.

\subsection{\label{subsec:Parameter studies} Parameter studies}
To assess the sensitivity of the proposed framework to variations in the number of measurement points and the SNR level of the noise added to the synthetically generated data, a series of parameter studies is conducted. As the melamine foam sample exhibits slightly lower estimation accuracy, all investigations are performed exclusively for this case, providing a more challenging and thus more informative evaluation scenario.
\begin{figure}[htb]
	\centering
	\includegraphics[width=0.95\columnwidth]{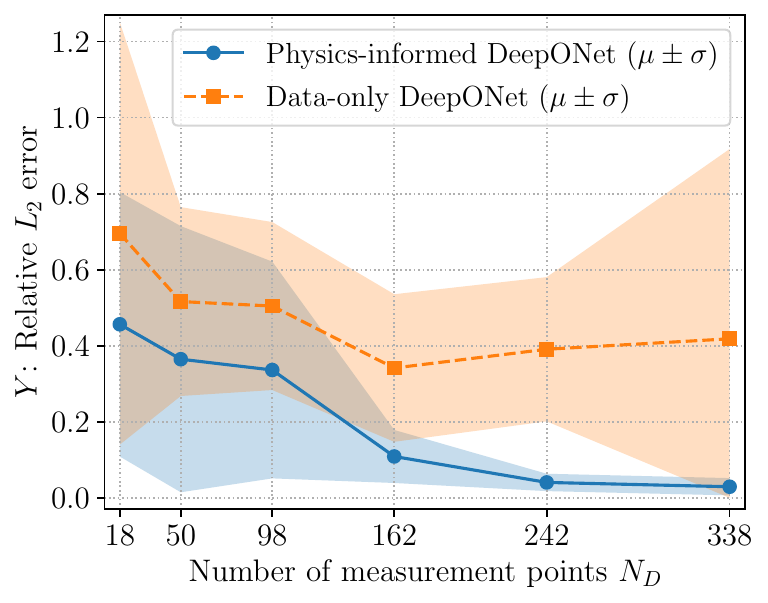}
	\caption{Mean relative $L_2$-norm error of the estimated surface admittance $Y$ as a function of the number of measurement points $N_D$, comparing the physics-informed DeepONet with a data-only DeepONet variant. The shaded region indicates the standard deviation $\sigma$ over the considered frequency range.}
	\label{fig:N_points_study}
\end{figure}
Figure~\ref{fig:N_points_study} illustrates the mean relative $L_2$-norm error of the estimated surface admittance $Y$ as a function of the number of measurement points $N_D$. The shaded regions represent the corresponding standard deviation $\sigma$, computed across all frequencies within the considered frequency range. The physics-informed DeepONet is compared to a data-driven variant in which the linearized momentum and Helmholtz losses are not enforced during training, evaluated at an SNR of \SI{40}{dB}. The physics-informed model consistently achieves lower errors across all spatial sampling densities and shows a clear convergence with increasing $N_D$, highlighting the enhanced accuracy and stability obtained through physics-based regularization as more measurement information becomes available. In contrast, the data-only approach exhibits significantly higher errors and no improvement with increasing data size, indicating limited robustness and generalization capability. The reduced standard deviation for the physics-informed approach, particularly for $N_D \geq 162$, further demonstrates its improved consistency across frequencies. Notably, even for sparse measurement configurations, the incorporation of physical constraints leads to a substantial gain in prediction accuracy. Overall, the results emphasize the critical role of physics-informed regularization in stabilizing the inverse problem and enabling reliable admittance estimation, especially under data-limited conditions.
\begin{figure}[htb]
	\centering
	\includegraphics[width=0.95\columnwidth]{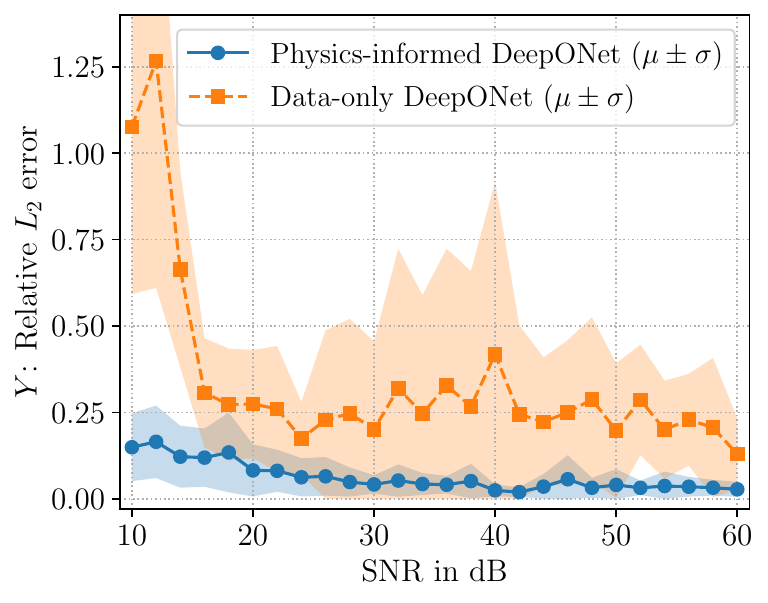}
	\caption{Mean relative $L_2$-norm error of the estimated surface admittance $Y$ as a function of the SNR level of the added noise, comparing the physics-informed DeepONet with a data-only variant. The shaded regions indicate the standard deviation $\sigma$ over the considered frequency range.}
	\label{fig:SNR_points_study}
\end{figure}

Fig.~\ref{fig:SNR_points_study} shows the relative $L_2$-norm error of the estimated surface admittance $Y$ as a function of the SNR level of the added noise, also comparing the physics-informed DeepONet with a purely data-driven variant. All results are obtained using $N_D = 338$ measurement positions. The physics-informed model demonstrates consistently low error levels across all SNR values, with a maximum mean error of 0.165 at \SI{12}{dB}, highlighting its robustness even under severe noise conditions. Its performance improves with increasing SNR up to approximately \SI{40}{dB}, beyond which the accuracy saturates. Concurrently, the standard deviation decreases, indicating enhanced robustness and consistency across frequencies. In contrast, the data-only model exhibits significantly higher errors and standard deviations, particularly at very low SNR levels, where reliable estimation is no longer achieved. Although its performance improves with increasing SNR, it remains substantially less accurate than the physics-informed approach across all noise levels. These findings highlight that incorporating physical constraints effectively regularizes the inverse problem and mitigates the impact of measurement noise, enabling stable and accurate admittance estimation even in challenging measurement scenarios. This constitutes a key advantage over conventional methods, in which measurement noise amplifies the ill-posedness of the inverse problem and degrades solution quality.

\section{\label{sec:Conclusions}Conclusions and outlook}
This work introduces a physics-informed neural operator framework for \textit{in situ} estimation of the frequency-dependent acoustic surface admittance from near-field measurements of sound pressure and normal particle velocity. In contrast to conventional approaches, which depend on explicit wave propagation models or computationally intensive iterative inversion procedures, the proposed method formulates the admittance estimation problem as a physics-constrained operator learning task. This enables the simultaneous inference of acoustic field quantities and the surface admittance, without requiring an explicit wave-propagation or forward model. By incorporating frequency as an input variable, the approach enables the prediction of the complete admittance spectrum in a single training process, thereby eliminating the need for frequency-wise inversion. Physical consistency is enforced by embedding the governing acoustic relations, including the Helmholtz equation, the linearized momentum equation, and the Robin boundary condition, directly into the training objective. This allows the admittance to be identified as a globally consistent parameter that satisfies both the measurement data and the underlying physics. The results obtained for synthetically generated data of a finite planar sample with two distinct surface admittance spectra, derived from impedance tube measurements of two porous materials, demonstrate accurate estimation of both the dissipative and reactive components of the surface admittance over a broad frequency range. The proposed approach achieves consistently low estimation errors while preserving a high degree of spatial consistency with the reference solutions. In addition, the model provides continuous and physically consistent predictions of pressure and particle velocity at unseen spatial locations. As expected, reduced accuracy is observed in frequency regimes with weak boundary sensitivity or increasingly complex spatial field structures. Parameter studies further show that the physics-informed formulation significantly enhances robustness compared to purely data-driven approaches, particularly under spatially sparse and noisy measurement conditions.

Despite these promising results, the method relies on a comparatively large amount of high-quality training data. However, such data can be feasibly acquired using modern measurement techniques, for instance automated robotic scanning systems. Moreover, neural network model interpretability is limited compared to analytical methods, and careful hyperparameter optimization is required for stable training.

Future work will focus on experimental validation with real measurement data, extension to more challenging non-free-field environments such as realistic rooms with pronounced reflections and modal behavior, and systematic studies of different sound-incidence conditions. Overall, the proposed framework establishes a promising neural network-based approach for \textit{in situ} acoustic material characterization, achieving physically consistent predictions without relying on an explicitly defined wave propagation model.

\begin{acknowledgments}
The authors would like to thank their colleagues from the material characterization research group for the valuable discussions. This work has received funding from the European Union’s Horizon Europe research and innovation program under the project "MELCHIOR", grant agreement No. 101073899.
\end{acknowledgments}

\section*{Author declarations}
\subsection*{Conflict of interest}
The authors have no conflict of interest to declare.

\section*{Data availability}
The data that support the findings of this study are available from the corresponding author upon reasonable request.

\bibliography{bibliography}

\end{document}